\documentclass[sigconf]{acmart} %

\usepackage{multirow}
\usepackage{xspace}
\usepackage[normalem]{ulem}

\newcommand{\ours}{\textsc{Octet}\xspace}
\newcommand{\taxos}{online catalog taxonomies\xspace}
\newcommand{\taxo}{online catalog taxonomy\xspace}
\newcommand{\gnnAware}{structural\xspace}
\newcommand{\GnnAware}{Structural\xspace}

\newcommand{\ie}{\textit{i.e.}} %
\newcommand{\eg}{\textit{e.g.}} %
\newcommand{\start}[1]{\vspace{0.3mm}\noindent{{\bf #1}}}

\usepackage{amsmath,amsfonts,bm}

\newcommand{\figleft}{{\em (Left)}}

\newcommand{\figright}{{\em (Right)}}

\def\eqref#1{equation~\ref{#1}}

\def\1{\bm{1}}

\def\vg{{\bm{g}}}
\def\vh{{\bm{h}}}

\def\vv{{\bm{v}}}
\def\vw{{\bm{w}}}

\def\mH{{\bm{H}}}

\def\mL{{\bm{L}}}

\def\mR{{\bm{R}}}
\def\mS{{\bm{S}}}

\def\mW{{\bm{W}}}

\DeclareMathAlphabet{\mathsfit}{\encodingdefault}{\sfdefault}{m}{sl}
\SetMathAlphabet{\mathsfit}{bold}{\encodingdefault}{\sfdefault}{bx}{n}

\def\gG{{\mathcal{G}}}

\def\gM{{\mathcal{M}}}

\def\gR{{\mathcal{R}}}

\newcommand{\normlone}{L^1}
\newcommand{\normltwo}{L^2}

\DeclareMathOperator*{\argmax}{arg\,max}

\definecolor{gred}{RGB}{219,68,55}
\definecolor{gblue}{RGB}{66,133,244}
\definecolor{gyellow}{RGB}{244,180,0}
\definecolor{ggreen}{RGB}{15,157,88}
\definecolor{ggrey}{RGB}{115,115,115}

\newcommand{\colorR}[1]{\textcolor{gred}{\textbf{#1}}}
\newcommand{\colorG}[1]{\textcolor{ggreen}{\textbf{#1}}}
\newcommand{\colorB}[1]{\textcolor{gblue}{\textbf{#1}}}

\setcopyright{rightsretained}
\begin{document}
\copyrightyear{2020}
\acmYear{2020}

\acmConference[KDD '20]{Proceedings of the 26th ACM SIGKDD Conference on Knowledge Discovery and Data Mining}{August 23--27, 2020}{Virtual Event, CA, USA}
\acmBooktitle{Proceedings of the 26th ACM SIGKDD Conference on Knowledge Discovery and Data Mining (KDD '20), August 23--27, 2020, Virtual Event, CA, USA}
\acmDOI{10.1145/3394486.3403274}
\acmISBN{978-1-4503-7998-4/20/08}

\title{\ours: Online Catalog Taxonomy Enrichment with Self-Supervision}

\author{Yuning Mao$^{1}$, Tong Zhao$^{2}$, Andrey Kan$^{2}$, Chenwei Zhang$^2$, }
\author{Xin Luna Dong$^2$, Christos Faloutsos$^3$, Jiawei Han$^1$}
\affiliation{
\institution{$^1$University of Illinois at Urbana-Champaign $\quad$ $^2$Amazon $\quad$ $^3$Carnegie Mellon University}
\institution{$^1$\{yuningm2, hanj\}@illinois.edu $\quad$ $^2$\{zhaoton, avkan, cwzhang, lunadong\}@amazon.com $\quad$ $^3$christos@cs.cmu.edu}
}

\renewcommand{\shortauthors}{Yuning Mao, et al.}

\begin{abstract}
  Taxonomies have found wide applications in various domains, especially online for item categorization, browsing, and search.
  Despite the prevalent use of \taxos, most of them in practice are maintained by humans, which is labor-intensive and difficult to scale.
  While \textit{taxonomy construction from scratch} is considerably studied in the literature, how to effectively enrich existing incomplete taxonomies remains an open yet important research question. Taxonomy enrichment not only requires the robustness to deal with emerging terms but also the consistency between existing taxonomy structure and new term attachment.
  In this paper, we present a self-supervised end-to-end framework, \ours, for Online Catalog Taxonomy EnrichmenT. 
  \ours leverages heterogeneous information unique to \taxos such as user queries, items, and their relations to the taxonomy nodes while requiring no other supervision than the existing taxonomies. 
  We propose to distantly train a sequence labeling model for term extraction and employ graph neural networks (GNNs) to capture the taxonomy structure as well as the query-item-taxonomy interactions for term attachment.
  Extensive experiments in different online domains demonstrate the superiority of \ours over state-of-the-art methods via both automatic and human evaluations. Notably, \ours enriches an \taxo in production to \textit{2 times larger} in the open-world evaluation.
\end{abstract}

\maketitle

\section{Introduction}
\begin{figure}[t]
    \includegraphics[width=1.0\linewidth]{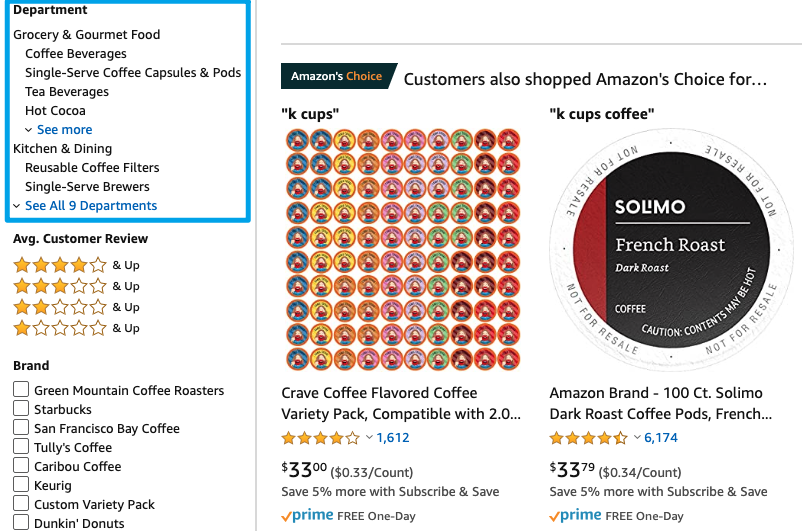}
    \vspace*{-.55cm}
    \caption{The most relevant taxonomy nodes are shown on the left when a user searches ``k cups'' on Amazon.com. }
    \label{fig:coffee_department}
    \vspace*{-.5cm}
\end{figure}

Taxonomies, the tree-structured hierarchies that represent the hypernymy (Is-A) relations, have been widely used in different domains, such as information extraction~\cite{Demeester2016LiftedRI}, question answering~\cite{Yang2017EfficientlyAT}, and recommender systems~\cite{huang2019taxonomy}, for the organization of concepts and instances as well as the injection of structured knowledge in downstream tasks.
In particular, \taxos serve as a building block of e-commerce websites (\eg, Amazon.com) and business directory services (\eg, Yelp.com) for both customer-facing and internal applications, such as query understanding, item categorization \cite{mao-etal-2019-hierarchical}, browsing, recommendation \cite{huang2019taxonomy}, and search \cite{wang2014hierarchical}.

Fig.~\ref{fig:coffee_department} shows one real-world example of how the product taxonomy at Amazon.com is used to facilitate online shopping experience:
when a user searches ``k cups'', the most relevant nodes (types) in the taxonomy \textit{Grocery \& Gourmet Food} are shown on the left sidebar. 
The taxonomy here serves multiple purposes. First, the user can browse relevant nodes to refine the search space if she is looking for a more general or specific type of items (\eg, ``Coffee Beverages'').
Second, the taxonomy benefits query understanding by identifying that ``k cups'' belongs to the taxonomy \textit{Grocery \& Gourmet Food} and mapping the user query ``k cups'' to the corresponding taxonomy node ``Single-Serve Coffee Capsules \& Pods''.
Third, the taxonomy allows query relaxation and makes more items searchable if the search results are sparse. For instance, not only ``Single-Serve Coffee Capsules \& Pods'' but also other coffee belonging to its parent type ``Coffee Beverages'' can be shown in the search results.

Despite the prevalent use and benefits of \taxos, most of them in practice are still built and maintained by human experts.
Such manual practice embodies knowledge from the experts but is meanwhile labor-intensive and difficult to scale. 
On Amazon.com, the taxonomies usually have thousands of nodes, not necessarily enough to cover the types of billions of items:
we sampled roughly 3 million items in Grocery domain on Amazon.com and found that over 70\% of items do not directly mention the types in the taxonomy, implying a mismatch between knowledge organization and item search.
As a result, automatic taxonomy construction has drawn significant attention.

Existing methods on taxonomy construction fail to work effectively on \taxos for the following reasons.
Most prior methods~\cite{kozareva2010semi,bansal2014structured,bordea2016semeval,shwartz2016improving,gupta2017taxonomy,mao2018end} are designed for taxonomy construction from general text corpora (\eg, Wikipedia), limiting their applicability to text-rich domains.
The ``documents'' in e-commerce (\eg, item titles), however, are much shorter and pose particular challenges.
First, it is implausible to extract terms with heuristic approaches~\cite{kozareva2010semi} from item titles and descriptions, since vendors can write them in arbitrary ways.
Second, it is highly unlikely to find co-occurrences of hypernym pairs in the item titles due to their conciseness, making Hearst patterns~\cite{hearst1992automatic,panchenko2016taxi} and dependency parse-based features~\cite{mao2018end} infeasible.
For instance, one may often see ``US'' and ``Seattle'' in the same document, but barely see ``Beverages'' and ``Coffee'' in the same item title.
Third, blindly leveraging the co-occurrence patterns could be misleading: in an item titled ``\textit{Triple Scoop Ice Cream Mix, Premium Strawberry}'', ``Strawberry'' and ``Ice Cream'' co-occur  but ``Strawberry'' is the flavor of the ``Ice Cream'' rather than its hypernym.
The situation worsens as \taxos are never static. There are new items (and thus new terms) emerging every day, making \textit{taxonomy construction from scratch} less favorable, since in practice we cannot afford to rebuild the whole taxonomy frequently and the downstream applications also require stable taxonomies to organize knowledge.

To tackle the above issues of \textit{taxonomy construction from scratch}, we target the \textit{taxonomy enrichment} problem, which discovers emerging concepts\footnote{We use ``concept'',  ``term'', ``type'', ``category'', and ``node'' interchangeably.} and attaches them to the existing taxonomy (named \textit{core taxonomy}) to precisely understand new customer interests.
Different from taxonomy construction from scratch, the core taxonomies, which are usually built and maintained by experts for quality control and actively used in production, provide both valuable guidance and restrictive challenges for taxonomy enrichment.
On the challenge side, the \textit{core taxonomy} requires term attachment to follow the existing taxonomy schema instead of arbitrarily building from scratch. 
On the bright side, we can base our work on the core taxonomy, which usually contains high-level and qualified concepts representing the fundamental categories in a domain (such as ``Beverages'' and ''Snacks`` for Grocery) and barely needs modification (or cannot be automatically organized due to business demands), but lacks fine-grained and emerging terms (such as ``Coconut Flour''). 
There are only a few prior works focused on taxonomy enrichment, which either employ simple rules~\cite{jurgens2015reserating} or represent taxonomy nodes by their associated items and totally neglect the lexical semantics of the concepts themselves~\cite{wang2014hierarchical}.
In addition, prior studies~\cite{jurgens2016semeval,schlichtkrull2016msejrku} require manual training data and fail to exploit the structural information of the existing taxonomy.

Despite the challenges, a unique opportunity for \taxo enrichment is the availability of rich user behavior logs: vendors often carefully choose words to describe the type of their items and associate the items with appropriate taxonomy nodes to get more exposure;
customers often include the item type in their queries 
and the majority of the clicked (purchased) items are instances of the type they are looking for. 
Such interactions among queries, items, and taxonomy nodes offer distinctive signals for hypernymy detection, which is unavailable in general-purpose text corpora. 
For instance, if a query mentioning ``hibiscus tea'' leads to the clicks of items associated with taxonomy node ``herbal tea'' or ``tea'', we can safely infer strong connections among ``hibiscus tea'', ``herbal tea'', and ``tea''.
Existing works~\cite{wang2014hierarchical,liu2019user}, however, only utilize the user behavior heuristically to extract new terms or reduce the prediction space of hypernymy detection.

\begin{figure*}[ht]
    \includegraphics[width=1.01\linewidth]{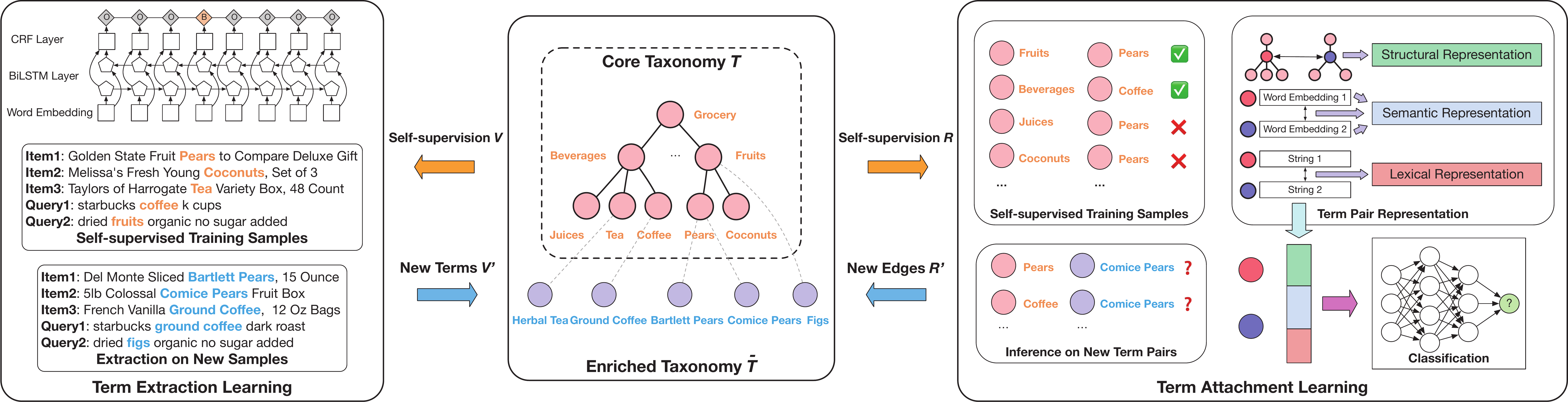}
    \vspace*{-.3cm}
    \caption{An overview of the proposed framework \ours. Item profiles and user queries in the target domain serve as framework input and the core taxonomy is used as self-supervision. New terms are automatically extracted via sequence labeling. Heterogeneous sources of signals, including the structure of the core taxonomy, the query-item-taxonomy interactions, and the lexical semantics of the term pairs are leveraged for hypernymy detection during term attachment.}
    \label{fig:framework}
    \vspace*{-.35cm}
\end{figure*}

In this paper, we present a self-supervised end-to-end framework, \ours, for \taxo enrichment.
\ours\ is novel in three aspects. First, \ours\ identifies new terms from item titles and user queries; it employs a sequence labeling model that is shown to significantly outperform typical term extraction methods. Second, to tackle the lack of text corpora, \ours leverages heterogeneous sources of signals; it captures the \textit{lexical semantics} of the terms and employs Graph Neural Networks (GNNs) to model the \textit{structure of the core taxonomy} as well as the \textit{query-item-taxonomy interactions} in user behavior. Third, \ours requires no human effort for generating training labels as it uses the core taxonomy for self-supervision during both term extraction and term attachment. 
We conduct extensive experiments on real-world \taxos to verify the effectiveness of \ours via automatic, expert, and crowdsourcing evaluations.
Experimental results show that \ours outperforms state-of-the-art methods by 46.2\% for term extraction and 11.5\% for term attachment on average.
Notably, \ours doubles the size (2,163 to 4,355 terms) of an \taxo in production with 0.881 precision.

\start{Contributions.}
(1) We introduce a self-supervised end-to-end framework, \ours, for \taxo enrichment; 
\ours automatically extracts emerging terms and attaches them to the core taxonomy of a target domain with no human effort.
(2) We propose a GNN-based model that leverages heterogeneous sources of information, especially the structure of the core taxonomy and query-item-taxonomy interactions, for term attachment.
(3) Our extensive experiments show that \ours significantly improves over state-of-the-art methods under automatic and human evaluations.

\section{Task Formulation}

\start{Notations.} 
We define a taxonomy $T = (V, R)$ as a tree-structured hierarchy with term set $V$ and edge set $R$.
A term $v \in V$ can be either single-word or multi-word (\eg, ``Yogurt'' and ``Herbal Tea'').
The edge set $R$ indicates the Is-A relationship between $V$ (hypernym pairs such as ``Coffee'' -> ``Ground Coffee'').
The \taxos, which can be found in almost all online shopping websites such as Amazon.com and eBay, and business directory services like Yelp.com, maintain the hypernym relationship of their items (\eg, products or businesses).
We define a \textit{core taxonomy} as a pre-given partial taxonomy that is usually manually curated and stores the high-level concepts in the target domain.
We denote user behavior logs as $B = (Q, I)$, which record the user queries $Q$ in a search engine and corresponding clicked items $I$.
The items $I$ are represented by item profiles such as titles and descriptions. $I$ is associated with (assigned to) nodes $V$ according to their types by item categorization (done by vendors or algorithms).

\start{Problem Definition.} Let $T = (V, R)$ be a core taxonomy and $B = (Q, I)$ be user behavior logs, the {\em taxonomy enrichment} problem extends $T$ to $\bar{T} = (\bar{V}, \bar{R})$ with $\bar{V} = V \cup V'$, $\bar{R} = R \cup R'$, where $V'$ contains new terms extracted from $Q$ and $I$, and $R'$ contains pairs $(v, v'), v\in V, v' \in V'$, representing that $v$ is a hypernym of $v'$. 

\start{Non-Goals.}
Although \ours works for terms regardless of their granularity, we keep $T$ unchanged as in~\cite{wang2014hierarchical,jurgens2016semeval} since we would like to keep the high-level expert-curated hypernym pairs intact and focus on discovering fine-grained terms.
Following convention~\cite{jurgens2016semeval}, we do not identify the hypernym relationship between newly discovered types ($(v_1', v_2')$, $v_1', v_2' \in V'$).
\ours already makes an important first step to solve the problem as our analysis shows that over 80\% terms are leaf nodes in the core taxonomy.

\section{The \ours Framework}
In this section, we first give a framework overview of the learning goal of term extraction and term attachment.
We then elaborate on how to employ self-supervision of the core taxonomy to conduct term extraction via sequence labeling (Sec.~\ref{sec:term_extract}) and term attachment via GNN-based heterogeneous representation (Sec.~\ref{sec:term_attach}).

\subsection{Framework Overview}
\ours consists of two inter-dependent stages: \textbf{term extraction} and \textbf{term attachment}, which, in a nutshell, solves the problem of ``which terms to attach'' and ``where to attach'', respectively.
Formally, the term extraction stage extracts the new terms $V'$ (with the guidance from $V$) that are to be used for enriching $T$ to $\bar{T}$.
The term attachment stage takes $T$ and $V'$ as well as the sources of $V'$ (\ie, $Q$ and $I$) as input and attaches each new term $v' \in V'$ to a term $v \in V$ in $T$, forming the new edge set $R'$.
\ours is readily deployable to different domains with no human effort, \ie, with no additional resources other than $T$, $Q$, $I$, and their interactions.

Fig.~\ref{fig:framework} shows an overview of \ours. 
At a high level, we regard term extraction as a sequence labeling task and employ a sequence labeling model with distant supervision from $V$ to extract new terms $V'$.
We show in Sec.~\ref{sec:term_extract_exp} that such a formulation is beneficial for the term extraction in \taxo enrichment.
For term attachment, existing hypernym pairs $R$ on $T$ are used as self-supervision.
Structure of the core taxonomy as well as interactions among queries, items, and taxonomy nodes are captured via graph neural networks (GNNs) for \gnnAware representation learning.
Meanwhile, the lexical semantics of the terms is employed and provides complementary signals for hypernymy detection (Sec.~\ref{sec:feat_ablation}).
Each new term $v' \in V'$ is attached to one existing term $v \in V$ on $T$ based on the term pair representation.

\subsection{Term Extraction Learning}\label{sec:term_extract}
Term extraction extends $T=(V, R)$ with new terms $V'$.
Extracting new terms from item profiles $I$ and user queries $Q$ has two benefits.
First, they are closely related to the dynamics of customer needs, which is essential for the enrichment of user-oriented \taxos and their deployment in production. Second, extraction from $I$ and $Q$ naturally connects type terms to user behavior, preparing rich signals required for generating \gnnAware term representations during term attachment.

We propose to treat \textit{term extraction for \taxo enrichment} as a sequence labeling task.
Tokens in the text are labeled with the BIOE schema where each tag represents the beginning, inside, outside, and ending of a chunk, respectively. 
Instead of collecting expensive human annotations for training~\cite{zheng2018opentag}, we propose to adopt distant supervision by using the existing terms $V$ as self-supervised labels.
Specifically, we find in $Q$ and $I$ the mentions of $V$ and label those mentions as the desired terms to be extracted. For example, item ``Golden State Fruit Pears to Compare Deluxe Gift'' with associated taxonomy node ``Pears'' will be labeled as "O O O B O O O O" for model learning.
This approach has several advantages. First, unlike unsupervised term extraction methods~\cite{kozareva2010semi}, we train a sequence labeling model to ensure that only the terms in the same domain as the existing terms $V$ are extracted.
Second, sequence labeling is more likely to extract terms of the desired type while filtering terms of undesired types (such as the brand or flavor of items) by inspecting the sentence context, which typical context-agnostic term extraction methods~\cite{liu2015mining,shang2018automated} fail to do.

We train a BiLSTM-CRF model~\cite{panchendrarajan2018bidirectional} to fulfill term extraction in \ours.
Briefly speaking, each token in the text is represented by its word embedding, fed into a bidirectional LSTM layer for contextualized representation. 
Then, tag prediction is conducted by a conditional random field (CRF) layer operating on the representation of the current token and the previous tag. 
A similar model architecture can be found in~\citet{zheng2018opentag}.

Note that distant supervision may inevitably introduce noises in the labels.
Nevertheless, we show in Sec.~\ref{sec:term_extract_exp} that \ours obtains superior performance in term extraction (precision@100=0.91). %
In addition, \ours is likely to have low confidence in attaching incorrectly extracted terms, allowing further filtering with threshold setting during term attachment (Sec.~\ref{sec:tradeoff}).

\subsection{Term Attachment Learning}\label{sec:term_attach}
Term attachment extends taxonomy $(\bar{V} = V \cup V', R)$ with new edges $R'$ between terms in $V$ and $V'$.
Following common practice~\cite{wang2014hierarchical,bansal2014structured}, we consider a taxonomy $\bar{T}$ as a Bayesian network where each node $v \in \bar{V}$ is a random variable.
The joint probability distribution of nodes $\bar{V}$ can be then formulated as follows.
\begin{equation*}
    P(\bar{V} \mid \bar{T}, \Theta) = P(v_0) \prod_{v \in \bar{V} \setminus v_0}  P(v \mid p(v), \Theta),
\end{equation*}
where $v_0$ denotes the root node, $p(v)$ denotes the direct hypernym (parent) of node $v$, and $\Theta$ denotes model parameters.
Maximizing the likelihood with respect to the taxonomy structure $\bar{T}$ gives the optimal taxonomy $\bar{T}^*$. 
As we do not modify the structure of core taxonomy $T$, the formulation can be simplified as follows.
\begin{equation*}
\vspace*{-.1cm}
\begin{split}
    \bar{T}^* = \argmax_{\bar{T}} P(\bar{V} \mid \bar{T}, \Theta) 
    &= \argmax_{\bar{T}} \prod_{v \in \bar{V} \setminus v_0} P(v \mid p(v), \Theta) \\
    &= \argmax_{\bar{T}}\ \prod_{v' \in V'} P(v' \mid p(v'), \Theta).
\end{split}
\vspace*{-.1cm}
\end{equation*}
The problem definition further ensures that $p(v') \in V$ always holds true and thus no inter-dependencies exist between the new terms $v' \in V'$.
Therefore, we can naturally regard term attachment as a \textit{multi-class classification problem} according to $P(v' \mid p(v'), \Theta)$ where each $p(v') = v \in V$ is a class. 

One unique challenge for \taxo enrichment is the lack of conventional corpora.
For example, it is rare to find co-occurrences of multiple item types in a single query (<1\% in Grocery domain on Amazon.com), let alone the ``such as''-style patterns in Hearst-based methods~\cite{kozareva2010semi}.
Instead of using the limited text directly, we introduce signals from user behavior logs and the structure of the core taxonomy by modeling them in a graph (Sec.~\ref{sec:gnn}). The lexical semantics of the term pairs are further considered to better identify hypernymy relations (Sec.~\ref{sec:semantic_rep} \& \ref{sec:lexical_rep}).

\subsubsection{\GnnAware Representation}\hfill
\label{sec:gnn}

\noindent There is rich \gnnAware information for \taxo enrichment which comes from two sources. First, the neighborhood (\eg, parent and siblings) of a node $v \in V$ on $T$ can serve as a meaningful supplement for the semantics of $v$.
For example, one may not have enough knowledge of node $v=$ ``Makgeolli'' (Korean rice wine); but if she perceives that ``Sake'' (Japanese rice wine) is $v$'s  sibling and ``Wine'' is $v$'s parent, she would have more confidence in considering ``Makgeolli'' as one type of ``Alcoholic Beverages''.
Second, there exist abundant user behavior data, providing even richer structural information than that offered by the core taxonomy $T$.
Specifically, items $I$ are associated with the existing taxonomy nodes $V$.
New terms $V'$ are related to items $I$ and user queries $Q$ since $V'$ are extracted from the two sources.
Furthermore, $I$ and $Q$ are also connected via clicks.
Based on the observations above, we propose to learn a \gnnAware term pair representation to capture the structure in the core taxonomy and the query-item-taxonomy interactions as follows.

\start{Graph Construction.}
We construct a graph $\gG$ where the nodes consist of the existing terms $v \in V$ and new terms $v' \in V'$.
There are two sets of edges: one set is the same as $R$ in the core taxonomy $T$,
which captures the ontological structure in $T$.
The other set leverages the query-item-taxonomy interactions: for each new term $v' \in V'$, we find the user queries $Q_{v'}$ that mention $v'$ and collect the clicked items $I_{Q_{v'}}$ in the queries $Q_{v'}$. Then, we find the taxonomy nodes $\{v_i\}$ that $I_{Q_{v'}}$ is associated with. Finally, we add an edge between each $(v_i, v')$ pair.
For instance, when determining the parent of a new term ``Figs'', we find that some queries mentioning ``Figs'' lead to clicked items associated with the taxonomy node ``Fruits'', evincing strong relations between the two terms.

\start{Graph Embedding.}
We leverage graph neural networks (GNNs) to aggregate the neighborhood information in $\gG$ when measuring the relationship between a term pair.
Specifically, we take the rationales in relational graph convolutional networks (RGCNs)~\cite{kipf2016semi,schlichtkrull2018modeling}.
Let $\gR$ denote the set of relations, including ($r_1$) The neighbors of $v$ on the core taxonomy $T$. The neighbors can be the (grand)parents or children of $v$ and we compare different design choices in Sec.~\ref{sec:graph_ablation};
($r_2$) the interactions between $v \in V$ and $v' \in V'$ discovered in user behavior. We confine the interactions to be unidirectional (from $v$ to $v'$) since the terms $v \in V$ are already augmented with $r_1$ while there might be noise in the user behavior;
and ($r_3$) the self-loop of node $v$. The self-loop of $v$ ensures that the information of $v$ itself is preserved and those isolated nodes without any connections can still be updated using its own representation.

Let $N(v, r)$ denote the neighbors of node $v$ with relation $r$, and $\vh^{\text{0}}_v$ the initial input representation of node $v$. 
The hidden representation of $v$ at layer (hop) $l$ is updated as follows:
 $\vh^{l + 1}_v = \text{ReLU} (\sum_{r \in \gR} \sum_{i \in \{N(v, r)\}} \frac{1}{c_{v, r}} \mW^l_r \vh^{l}_i)$,
where $\mW^l_r$ is the matrix at layer $l$ for linear projection of relation $r$ and $c_{v, r}$ is a normalization constant.
We take the final hidden representation of each node $\vh^{L}_v$ (\textit{denoted as $\vg_v$}) as the graph embedding.

\start{Relationship Measure.}
\label{sec:sim_measure}
One straightforward way of utilizing the \gnnAware representation is to use the graph embeddings $\vg_v$ and $\vg_{v'}$ as the representation of term $v$ and $v'$.
Instead, we propose to measure the relationship between a term pair explicitly by cosine similarity and norm distance (more details in App.~\ref{app:repro}). 
We denote the relationship measure between two embeddings as $s(\vv, \vv')$.
One benefit of using $s(\vg_v, \vg_{v'})$ is that empirically we observe $s(\vg_v, \vg_{v'})$ alleviates overfitting compared with directly using $\vg_v$ and $\vg_v'$.
Also, using $s(\vg_v, \vg_{v'})$ makes the final output size much smaller and reduces the number of parameters in the following layer significantly.

\subsubsection{Semantic Representation}\hfill

\label{sec:semantic_rep}

\noindent We use the word embedding $\vw_{v}$ of each term $v$ to capture its semantics. 
For grouping nodes consisting of several item types (\eg, ``Fresh Flowers \& Live Indoor Plants'') and multi-word terms with qualifiers (\eg, ``Fresh Cut Bell Pepper''), we employ dependency parsing to find the noun chunks (``Fresh Flowers'', ``Live Indoor Plants'', and ``Fresh Cut Bell Pepper'') and respective head words (``Flowers'', ``Plants'', and ``Pepper'').
Intuitively, $(v, v')$ tends to be related as long as one head word of $v$ is similar to that of $v'$.
We thus use the relationship measure $s(\vv, \vv')$ defined in Sec.~\ref{sec:sim_measure} to measure the semantic relationship of the head words: $\mH(v, v') = \max_{i, j} s (\vw_{\text{head}^i(v)}, \vw_{\text{head}^{j}(v')})$,
where $\vw_{\text{head}^i(v)}$ denotes the word embedding of the i-th head word in the term $v$.
Finally, the overall semantic representation $\mS(v, v')$ is defined as $\mS(v, v') = [\mH(v, v'), \vw_v, \vw_{v'}]$, where ``,'' denotes the concatenation operation.

\subsubsection{Lexical Representation}\hfill

\label{sec:lexical_rep}
\noindent String-level measures prove to be very effective in hypernymy detection~\cite{bansal2014structured,bordea2016semeval,gupta2017taxonomy,mao2018end}.
For \taxos, we also find many cases where lexical similarity provides strong signals for hypernymy identification. 
For example, ``Black Tea'' is a hyponym of ``Tea'' and ``Fresh Packaged Green Peppers'' is a sibling of ``Fresh Packaged Orange Peppers''.
Therefore, we take the following lexical features~\cite{mao2018end} to measure the lexical relationship between term pairs: \emph{Ends with}, \emph{Contains}, \emph{Suffix match}, \emph{Longest common substring}, \emph{Length difference}, and \emph{Edit distance}.
Values in each feature are binned by range and each bin is mapped to a randomly initialized embedding, which would be updated during model training.
We denote the set of lexical features by $\gM$ and compute lexical representation as the concatenation of the lexical features: $\mL(v, v') = [\mL_i(v, v')]_{ i \in \gM }$.

\subsubsection{Heterogeneous Representation}\hfill

\noindent For each term pair $(v, v')$, we generate a heterogeneous term pair representation $\mR(v, v')$ by combining the representations detailed above.
$\mR(v, v')$ captures several orthogonal aspects of the term pair relationship, which contribute to hypernymy detection in a complementary manner (Sec.~\ref{sec:feat_ablation}). 
To summarize, the \gnnAware representation models the core taxonomy structure as well as underlying query-item-taxonomy interactions, whilst the semantic and lexical representations capture the distributional and surface information of the term pairs, respectively.
We further calculate $s(\vv, \vv')$ between the graph embedding $\vg_v$ ($\vg_v'$) of one term and the word embedding $\vw_{v'}$ ($\vw_{v}$ ) of the other term in the term pair, which measures the relationship of the term pair in different forms and manifests improved performance.
Formally, the heterogeneous term pair representation $\mR(v, v')$ is defined as follows.
\begin{equation*}
    \mR(v, v')\! =\! [s(\vg_v, \vg_{v'}), s(\vw_v, \vg_{v'}), s(\vg_v, \vw_{v'}), \mS(v, v'), \mL(v, v')].
\end{equation*}

\subsubsection{Model Training and Inference}\hfill

\noindent Similar to prior works~\cite{shwartz2016improving,mao2018end}, we feed $\mR(v, v')$ into a two-layer feed-forward network and use the output after the sigmoid function as the probability of hypernym relationship between $v$ and $v'$.
To train the term attachment module, we permute all the term pairs $(v_i, v_j)$ in $V$ as training samples and utilize the existing hypernym pairs $R$ on $T$ for self-supervision -- the pairs in $R$ are regarded as positive samples and other pairs are negative.
The heterogeneous term pair representation, including the structural representation, is learned in an end-to-end fashion.
We use binary cross-entropy loss as the objective function due to the classification formulation.
An alternative formulation is to treat term attachment as a hierarchical classification problem where the positive labels are all the ancestors of the term (instead of only its parent).
We found, however, that hierarchical classification does not outperform standard classification and thus opt for the simpler formulation following~\cite{wang2014hierarchical,bansal2014structured}.
For inference, we choose $v_i \in V$ with the highest probability among all the permuted term pairs $(v_i, v')$ as the predicted hypernym of $v'$.

\section{Experiments}
In this section, we examine the effectiveness of \ours via both automatic and human evaluations.
We first conduct experiments on term extraction (Sec.~\ref{sec:term_extract_exp}) and term attachment (Sec.~\ref{sec:term_attach_exp}) individually, and then perform an end-to-end open-world evaluation for \ours (Sec.~\ref{sec:e2e_exp}). Finally, we carefully analyze the performance of \ours via framework ablation and case studies (Sec.~\ref{sec:analysis}).

\subsection{Evaluation on Term Extraction}
\label{sec:term_extract_exp}

\subsubsection{Evaluation Setup.}\hfill
\label{sec:term_extract_setup}

\noindent We take the \textit{Grocery \& Gourmet Food} taxonomy (2,163 terms) on Amazon.com as the major testbed for term extraction.
We design three different evaluation setups with closed-world or open-world assumption as follows.

\start{Closed-world Evaluation.}
We first conduct a closed-world evaluation that  holds out a number of terms on the core taxonomy $T$ as the virtual $V'$.
In this way, we can ensure that the test set follows a similar distribution as the training set and the evaluation can be done automatically.
Specifically, we match the terms on $T$ with the titles of active items belonging to \textit{Grocery \& Gourmet Food} (2,995,345 in total).\footnote{We also observed positive results on term extraction from user queries at Amazon.com but omit the results in the interest of space.}
948 of the 2,163 terms are mentioned in at least one item title and 948,897 (item title, term) pairs are collected.
We split the pairs by the 948 matched terms into training set $V$ and test set $V'$ with a ratio of 80\% / 20\% and evaluate term recall on the unseen set $V'$.
Splitting the pairs by terms (instead of items) ensures that $V$ and $V'$ have no overlap, which is much more challenging than typical named entity recognition (NER) tasks but resembles the real-world use cases for \taxo enrichment.

\start{Open-world Evaluation.}
Open-world evaluation tests on new terms that do not currently exist in $T$, which is preferable since it evaluates term extraction methods in the same scenario as in production.
The downside is that it requires manual annotations as there are no labels for the newly extracted terms.
Therefore, we ask experts and workers on Amazon Mechanical Turk (MTurk) to verify the quality of the new terms.
As we would like to take new terms with high confidence for term attachment, we ask our taxonomists to measure the precision of top-ranked terms that each method is most confident at. The terms that are already on the core taxonomy $T$ are excluded and the top 100 terms of each compared method are carefully verified.
To evaluate from the average customers' perspective, we sample 1,000 items and ask MTurk workers to extract the item types in the item titles.
Different from expert evaluation, items are used as the unit for evaluation rather than terms. Precision and recall, weighted by the votes of the workers, are measured. More details of crowdsourcing are provided in App.~\ref{app:crowdsourcing}.

\start{Baseline Methods.}
For the evaluation on term extraction, we compare with two approaches widely used for taxonomy construction, namely noun phrase (NP) chunking and AutoPhrase~\cite{shang2018automated}.
Pattern-based methods~\cite{liu2019user} and classification-based methods with simple n-gram and click count features~\cite{wang2014hierarchical} perform poorly in our preliminary experiments and are thus excluded.
More details and discussions of the baselines are provided in App.~\ref{app:baseline}.

\subsubsection{Evaluation Results}\hfill

\begin{figure}[t]
    \includegraphics[width=.96\linewidth]{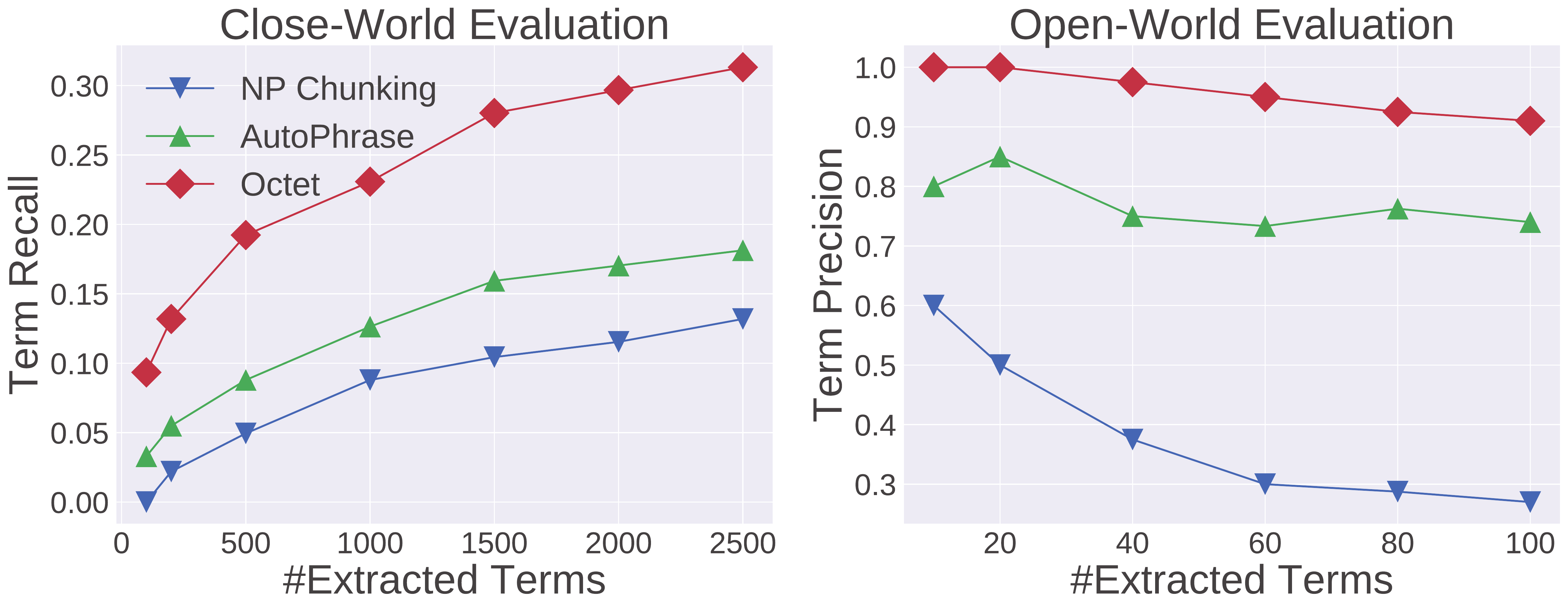}
    \vspace{-.3cm}
    \caption{Closed-world automatic evaluation on term recall and open-world expert evaluation on term precision.}
    \label{fig:term_eval}
    \vspace{-.25cm}
\end{figure}

\noindent For closed-world automatic evaluation, we calculate recall@K and show the comparison results in Fig.~\ref{fig:term_eval}~\figleft.
We observe that \ours consistently achieves the highest recall@K on the held-out test set.
The overall recall of all compared methods, however, is relatively low.
Nevertheless, we argue that the low recall is mainly due to the wording difference between the terms in the core taxonomy $T$ and item titles. 
As we will demonstrate in the open-world evaluation below, many extracted terms are valid but not on $T$, which also confirms that $T$ is very sparse and incomplete.

For open-world expert evaluation, we examine the terms each method is most confident at.
In AutoPhrase~\cite{shang2018automated}, each extracted term is associated with a confidence score.
In NP chunking and \ours, we use the frequency of extracted terms (instead of the raw frequency via string match) as their confidence score.
As shown in Fig.~\ref{fig:term_eval}~\figright, \ours achieves very high precision@K (0.91 when K=100), which indicates that the newly extracted terms not found on the core taxonomy are of high quality according to human experts and can be readily used for term attachment.
The compared methods, however, perform much worse on the top-ranked terms. In particular, the performance of NP chunking degenerates very quickly and only 27 of its top 100 extracted terms are valid.

We further show examples of the extracted terms in Table~\ref{table:term_example}. 
As one can see, NP chunking extracts many terms that are either not item types (\eg, ``Penzeys Spices'' is a company and ``No Sugar'' is an attribute) or less specific (\eg, ``Whole Bean'' Coffee).
AutoPhrase extracts some terms that are not of the desired type.
For example, ``Honey Roasted'' and ``American Flag'' are indeed high-quality phrases that appear in the item titles but not valid item types.
In contrast, \ours achieves very high precision on the top-ranked terms while ensuring that the extracted terms are of the same type as existing terms on the core taxonomy, which empirically verifies the superiority of formulating term extraction for \taxo enrichment as sequence labeling.

\begin{table}[t]
    \caption{Examples of top-ranked terms extracted by each approach. Valid terms for item types are marked in bold.} 
\label{table:term_example}
    \vspace*{-.3cm}
    \centering
    \scalebox{.82}{
    \begin{tabular}{p{2.2cm}p{2.6cm}p{2.6cm}}
        \toprule
         \textbf{NP Chunking} & \textbf{AutoPhrase}~\cite{shang2018automated} & \textbf{\ours}\\
         \midrule
        \textbf{dark chocolate}, whole bean, penzeys spices, \textbf{extra virgin olive oil}, net wt, a hint, no sugar
        & \textbf{almond butter}, honey roasted, \textbf{hot cocoa}, \textbf{tonic water}, \textbf{brown sugar}, \textbf{curry paste}, american flag
        & \textbf{coconut flour}, \textbf{ground cinnamon}, \textbf{red tea}, \textbf{ground ginger}, \textbf{green peas}, sweet leaf, \textbf{coconut syrup}
 \\
        \bottomrule
    \end{tabular}
    }
    \vspace*{-.25cm}
\end{table}

Finally, we show the results of open-world crowdsourcing evaluation in Table~\ref{table:extract_mturk}.
\ours again achieves much higher precision and F1 score than the baseline methods.
AutoPhrase obtains higher recall as we found that it tends to extract multiple terms in each sample, whereas there is usually one item type.
The recall of all the compared methods is still relatively low, which is possibly due to the conciseness and noise in the item titles (recall examples in Fig.~\ref{fig:framework}) and leaves much space for future work.

\begin{table}[t]
    \caption{Performance comparison in the open-world crowdsourcing evaluation on 1,000 sampled items. \ours achieves significantly higher precision and best F1 score. } 
    \label{table:extract_mturk}
    \vspace*{-.3cm}
    \centering
    \scalebox{.95}{
    \begin{tabular}{l cc c}
        \toprule
         \textbf{Method} & \textbf{Precision} & \textbf{Recall} & \textbf{F1} \\
        \midrule
        NP Chunking & 12.3 & 20.4 & 15.4  \\
        AutoPhrase~\cite{shang2018automated} & 20.9 & \textbf{41.3} & 27.8  \\
        \ours & \textbf{87.5} & 24.9 & \textbf{38.8}  \\
        \bottomrule
    \end{tabular}
    }
    \vspace*{-.3cm}
\end{table}

\subsection{Evaluation on Term Attachment}
\label{sec:term_attach_exp}

\subsubsection{Evaluation Setup.}\hfill

\noindent For term attachment, we also conduct both closed-world and open-world evaluations, which involves ablating the core taxonomy $T$ and attaching newly extracted terms, respectively.
In contrast, most of prior studies~\cite{bansal2014structured,jurgens2016semeval,mao2018end} only perform closed-world evaluation.

\start{Closed-world Evaluation.}
We take four taxonomies actively used in production as the datasets: \textit{Grocery \& Gourmet Food}, \textit{Home \& Kitchen}, and \textit{Beauty \& Personal Care} taxonomies at Amazon.com, and the \textit{Business Categories} at Yelp.com.\footnote{The taxonomies are available online. We use five-month user behavior logs on Amazon.com for \gnnAware representation and do not leverage user behavior on Yelp.com due to accessibility.  See App.~\ref{app:repro} for more details on data availability.} 
\textit{Amazon Grocery \& Gourmet Food} is used for framework analysis unless otherwise stated.
As the taxonomies are used in different domains and constructed according to their own business needs, they exhibit a wide spectrum of characteristics and almost have no overlap. 
Considering the real-world use cases where fine-grained terms are missing, we hold out the leaf nodes as the new terms $V'$ to be attached as in~\cite{jurgens2016semeval}.
We split $V'$ into training, development, and test sets with a ratio of 64\% / 16\% / 20\%, respectively.
Detailed statistics of the datasets can be found in Table~\ref{table:taxo_stats}.
Note that if we regard term attachment as a classification problem, each class would have very few training samples (\eg, 1193 / 298 $\approx$ 4 in \textit{Amazon Grocery \& Gourmet Food}), which calls for a significant level of model generalizability.

\begin{table}[ht]
    \caption{Taxonomy statistics for term attachment.} 
    \centering
    \label{table:taxo_stats}
    \vspace*{-.3cm}
    \scalebox{.8}{
    \begin{tabular}{l cccc c}
        \toprule
         \textbf{Taxonomy} & $|\textbf{V}|$ & $|\textbf{V}'|$ & $|\textbf{V}'_{\text{Train}}|$ & $|\textbf{V}'_{\text{Dev}}|$ & $|\textbf{V}'_{\text{Test}}|$ \\
        \midrule
        Amazon Grocery \& Gourmet Food & 298 & 1,865 & 1,193 & 299 & 373  \\
        Amazon Home \& Kitchen & 338 & 1,410 & 902 & 226 & 282  \\
        Amazon Beauty \& Personal Care & 109 & 454 & 290 & 73 & 91  \\
        Yelp Business Categories & 84 & 920 & 588 & 148 & 184  \\
        \bottomrule
    \end{tabular}
    }
    \vspace*{-.2cm}
\end{table}

\start{Open-world Evaluation.}
 We conduct an open-world evaluation for term attachment on \textit{Amazon Grocery \& Gourmet Food}.
Specifically, we ask the taxonomists to identify valid new terms among those discovered in the term extraction stage with frequency greater than 20. 106 terms are thus labeled as valid.
We then ask the taxonomists to attach the 106 terms manually to the core taxonomy as ground truth and evaluate systems on these new terms with the same criteria as in the closed-world evaluation.

\start{Evaluation Metrics.} 
We use Edge-F1 and Ancestor-F1~\cite{mao2018end} to measure the performance of term attachment.
\textbf{Edge-F1} compares the predicted edge $(v, v')$ with the gold edge $(p(v'), v')$, \ie, whether $v = p(v')$. 
We use $P_{\text{Edge}}$, $R_{\text{Edge}}$, and $F1_{\text{Edge}}$ to denote the precision, recall, and F1-score, respectively.
In particular, $P_{\text{Edge}} = R_{\text{Edge}} = F1_{\text{Edge}}$ if the number of predicted edges is the same as gold edges, \ie, when all the terms in $V'$ are attached.
  \textbf{Ancestor-F1} is more relaxed than Edge-F1 as it compares all the term pairs $(v_\text{sys}, v')$ with $(v_\text{gold}, v')$, where $v_\text{sys}$ represents the terms along the system predicted path (ancestors) to the root node, and $v_\text{gold}$ denotes those terms on the gold path.
Similarly, we denote the ancesor-based metrics as $P_{\text{Ancestor}} = \frac{| v_{\text{sys}} \wedge v_{\text{gold}} |} {| v_{\text{sys}} |}$ and $R_{\text{Ancestor}} = \frac{| v_{\text{sys}} \wedge v_{\text{gold}} |} {| v_{\text{gold}} |}$.

\start{Baseline Methods.}
As discussed earlier, there are few existing methods for taxonomy enrichment.
MSejrKu~\cite{schlichtkrull2016msejrku} is the winning method in the SemEval-2016 Task 14~\cite{jurgens2016semeval}.
{HiDir}~\cite{wang2014hierarchical} is the state-of-the-art method for \taxo enrichment.
In addition, we compare with the substring method {Substr}~\cite{bordea2016semeval}, and {I2T} that finds one's hypernym by examining where its related items are assigned to.
Two na\"ive baselines are also tested to better understand the difficulty of the task following convention~\cite{bansal2014structured,jurgens2016semeval}, where \textit{Random} attaches $v' \in V'$ randomly to $T$ and \textit{Root} attaches every term to the root node of $T$.
More details of the baselines are in App.~\ref{app:baseline}.

\subsubsection{Evaluation Results}\hfill

\noindent We first evaluate different methods under the closed-world assumption (Tables~\ref{table:grocery_main} \& ~\ref{table:attachment_additional}).
We observe that the Edge-F1 of two na\"ive baselines is very low since there are hundreds of $v \in V$ as candidates.
The performance of I2T is similar to Substr but still far from satisfactory, implying that there might be noise in the matching between $V'$ and $I$, and the associations between $I$ and $V$.
HiDir~\cite{wang2014hierarchical} and MSejrKu~\cite{schlichtkrull2016msejrku} achieve better performance than other baselines, especially in Edge-F1, while \ours outperforms all the compared methods by a large margin on both development and test sets across all of the four domains.

\begin{table}[t]
    \caption{Comparison results of closed-world evaluation on \textit{Amazon Grocery \& Gourmet Food}.}
    \label{table:grocery_main}
    \vspace*{-.3cm}
    \centering
    \scalebox{.88}{
    \begin{tabular}{l cccc}
        \toprule
               \multirow{2}{*}{\textbf{Method}}   & \multicolumn{2}{c}{\textbf{Dev}}   & \multicolumn{2}{c}{\textbf{Test}}\\
          & Edge-F1 & Ancestor-F1 & Edge-F1 & Ancestor-F1\\
        \midrule
        Random~\cite{bansal2014structured,jurgens2016semeval} & 0.3 & 30.3 & 0.5 & 31.2 \\
        Root & 0.3 & 41.1 & 0 & 41.8 \\
        I2T & 10.0 & 50.5 & 9.9 & 50.7 \\
        Substr~\cite{bordea2016semeval} & 8.4 & 49.9 & 10.7 & 52.9 \\
        HiDir~\cite{wang2014hierarchical} & 42.5 & 66.8 & 40.5 & 66.4 \\
        MSejrKu~\cite{schlichtkrull2016msejrku} & 58.9 & 80.6 & 53.1 & 76.7 \\
        \ours & \textbf{64.9} & \textbf{85.2} & \textbf{62.5} & \textbf{84.2} \\
        \bottomrule
    \end{tabular}
    }
\vspace*{-.3cm}
\end{table}

\begin{table}[t]
    \caption{Closed-world evaluation in different domains. Only competitive baseline methods that perform well on \textit{Amazon Grocery \& Gourmet Food} are listed.} 
    \label{table:attachment_additional}
    \centering
    \vspace*{-.3cm}
    \scalebox{.8}{
    \begin{tabular}{ll cc}
        \toprule
        \textbf{Dataset} & \textbf{Method} & \textbf{Edge-F1} & \textbf{Ancestor-F1}\\
        \midrule
        \multirow{3}{*}{Amazon Home \& Kitchen}
        & HiDir~\cite{wang2014hierarchical} & 46.5 & 76.9 \\
        & MSejrKu~\cite{schlichtkrull2016msejrku}  & 46.0 & 74.1 \\
         &  \ours & \textbf{54.4} & \textbf{78.5} \\
         \midrule
        \multirow{3}{*}{Amazon Beauty \& Personal Care}
        & HiDir~\cite{wang2014hierarchical} & 34.1 & 71.8 \\
        & MSejrKu~\cite{schlichtkrull2016msejrku}  & 49.5 & 75.0 \\
         &  \ours & \textbf{50.6} & \textbf{77.1} \\
         \midrule
        \multirow{3}{*}{Yelp Business Categories}
        & HiDir~\cite{wang2014hierarchical} & 19.6 & 35.8 \\
        & MSejrKu~\cite{schlichtkrull2016msejrku}  & 29.9 & 35.7 \\
         &  \ours & \textbf{32.6} & \textbf{43.5} \\
        \bottomrule
    \end{tabular}
    }
\vspace*{-.2cm}
\end{table}

For the open-world evaluation, we compare \ours with the best performing baselines MSejrKu~\cite{schlichtkrull2016msejrku} and HiDir~\cite{wang2014hierarchical} on the expert-labeled data (Table~\ref{table:attachment_open}).
Perhaps unsurprisingly, the performance of each method is lower than that under the closed-world evaluation, since the distributions of the existing terms on $T$ and the newly extracted terms are largely different (shown in Sec.~\ref{sec:term_extract_exp}).
\ours again achieves the best performance thanks to its better generalizability.

\begin{table}[t]
    \caption{Open-world expert evaluation. Note that the seemingly low \textit{absolute} performance is comparable to results on other datasets \cite{bansal2014structured,mao2018end} due to the difficulty of the task.} 
    \label{table:attachment_open}
    \centering
    \vspace*{-.3cm}
    \scalebox{.95}{
    \begin{tabular}{l cc}
        \toprule
         \textbf{Method} & \textbf{Edge-F1} & \textbf{Ancestor-F1}\\
        \midrule
        HiDir~\cite{wang2014hierarchical} & 28.3 & 59.2 \\
        MSejrKu~\cite{schlichtkrull2016msejrku}  & 29.3 & 61.2 \\
        \ours & \textbf{30.2} & \textbf{67.5} \\
        \bottomrule
    \end{tabular}
    }
\vspace*{-.3cm}
\end{table}

\subsection{End-to-End Evaluation}
\label{sec:e2e_exp}

Besides the individual evaluations on term extraction and term attachment, we perform a novel \textit{end-to-end open-world evaluation} that helps us better understand the quality of the enriched taxonomy by examining errors throughout the entire framework: whether (A) the extracted term is invalid or (B) the term is valid but the attachment is inaccurate.
To our knowledge, such an end-to-end evaluation has not been conducted in prior studies.

We evaluate \ours on \textit{Amazon Grocery \& Gourmet Food} using Amazon MTurk.
The details of crowdsourcing can be found in App.~\ref{app:crowdsourcing}.
In total, \ours extracts and attaches 2,192 new terms from the item titles described in Sec.~\ref{sec:term_extract_setup}, doubling the size of the existing taxonomy (2,163 terms).
As listed in Table~\ref{table:end2end},  only 6.5\% of extracted terms are considered invalid by average customers (MTurk workers). 
The top-1 edge precision and ancestor precision are relatively low, but they are comparable to the state-of-the-art performance on similar datasets with \textit{clean} term vocabulary \cite{mao2018end}.
The neighbor precision, which considers the siblings on the predicted path as correct, is very high (88.1).
One can further improve the precision by filtering low-confidence terms or allowing top-k prediction (Sec.~\ref{sec:tradeoff}).

\begin{table}[h]
\vspace*{-.3cm}
    \caption{Open-world end-to-end evaluation for \ours.} 
    \label{table:end2end}
    \centering
    \vspace*{-.3cm}
    \scalebox{.8}{
    \begin{tabular}{cc ccc}
        \toprule
          \textbf{Error A\%} & \textbf{Error B\%} & \textbf{Edge Prec}& \textbf{Ancestor Prec}&\textbf{Neighbor Prec}\\
        \midrule
         6.5 & 11.9 & 22.1 & 40.9 & 88.1 \\
        \bottomrule
    \end{tabular}
    }
\vspace*{-.35cm}
\end{table}

\subsection{Framework Analysis}
\label{sec:analysis}

\subsubsection{Feature Ablation}
\label{sec:feat_ablation}

\noindent We analyze the contribution of each representation for the term pair (Table~\ref{table:ablation_feat}). 
As one can see, only using word embedding (W) does not suffice to identify hypernymy relations while adding the semantics of head words (H) explicitly boosts the performance significantly.
Lexical (L) features are very effective and combining lexical representation with semantic representation (L + W + H) brings 17.2 absolute improvement in Edge-F1.
The structural information is very useful in that even if we use word embedding (W) as the input of the \gnnAware representation (G), the Edge-F1 improves by 4.8x and the Ancestor-F1 improves by 59.3\% upon W.
The full model that incorporates various representations performs the best, indicating that they capture different aspects of the term pair relationship and are complementary to each other.

\begin{table}[t]
    \caption{Ablation study of word embedding (W), head word semantics (H), lexical representation (L),  and \gnnAware graph-based representation (G).} 
    \centering
    \label{table:ablation_feat}
        \vspace*{-.3cm}
    \scalebox{.85}{
    \begin{tabular}{l cccc}
        \toprule
               \multirow{2}{*}{\textbf{Representation}}   & \multicolumn{2}{c}{\textbf{Dev}}   & \multicolumn{2}{c}{\textbf{Test}}\\
          & Edge-F1 & Ancestor-F1 & Edge-F1 & Ancestor-F1\\
        \midrule
        W & 12.0 & 50.4 & 11.0 & 49.7 \\
        H & 30.8 & 62.9 & 29.2 & 64.4 \\
        W + H & 41.8 & 68.8 & 39.7 & 67.8 \\
        L & 48.2 & 74.7 & 42.6 & 70.3 \\
        L + W & 57.9 & 79.7 & 52.6 & 76.9 \\
        G & 57.5 & 80.3 & 50.9 & 78.6 \\
        L + W + G & 63.6 & 82.6 & 56.0 & 79.9 \\
        L + W + H & 62.5 & 83.6 & 59.8 & 81.6 \\
        L + W + H + G & \textbf{64.9} & \textbf{85.2} & \textbf{62.5} & \textbf{84.2} \\
        \bottomrule
    \end{tabular}
    }
    \vspace*{-.2cm}
\end{table}

\subsubsection{Graph Ablation}
\label{sec:graph_ablation}

\noindent We analyze different variants regarding the design choices in the \gnnAware representation.
As shown in Table~\ref{table:graph_choice}, considering multi-hop relations (\eg, grandparents and siblings) is better than only considering immediate family (\ie, parent and children).
The directionality of edges does not have a huge effect on the model performance, although the information of the ancestors tends to be more benefitial for Ancestor-F1 and descendants for Edge-F1.
Adding the query-item-taxonomy interactions in the user behavior ($r_2$) in addition to the structure in the core taxonomy further improves model performance, showing the benefits of leveraging user behavior for term attachment.

\begin{table}[t]
    \caption{Comparison of various design choices in the \gnnAware representation. C and P denote Child and Parent. $r_1$, $r_2$, $r_3$ denote structure in the core taxonomy, query-item-taxonomy interactions, and self-loop, respectively.} 
    \label{table:graph_choice}
    \centering
    \vspace*{-.3cm}
    \scalebox{.88}{
    \begin{tabular}{c l cc}
        \cmidrule[0.06em]{2-4}
         &\textbf{Design Choice} & \textbf{Edge-F1} & \textbf{Ancestor-F1}\\
        \cmidrule{2-4}
        \multirow{2}{*}{\rotatebox[origin=c]{90}{$L$}} &One-hop neighborhood & 50.4 & 75.9\\
        &Two-hop neighborhood & \textbf{60.1} & \textbf{83.0}\\
        \cmidrule{2-4}
        \multirow{3}{*}{\rotatebox[origin=c]{90}{$N(v, r_1)$}} &C->P & \textbf{60.1} & 83.0 \\
        &C<->P & 59.8 & 83.2 \\
        &P->C & 59.3 & \textbf{83.6} \\
        \cmidrule{2-4}
        \multirow{2}{*}{\rotatebox[origin=c]{90}{$\gR$}} &$\{r_1, r_3\}$ & 60.1 & 83.0\\
        &$\{r_1, r_2, r_3\}$ & \textbf{62.5} & \textbf{84.2}\\
        \cmidrule[0.06em]{2-4}
    \end{tabular}
    }
\vspace*{-.35cm}
\end{table}

\subsubsection{Performance Trade-Off}
\label{sec:tradeoff}

\begin{figure}[t]
    \includegraphics[width=1.01\linewidth]{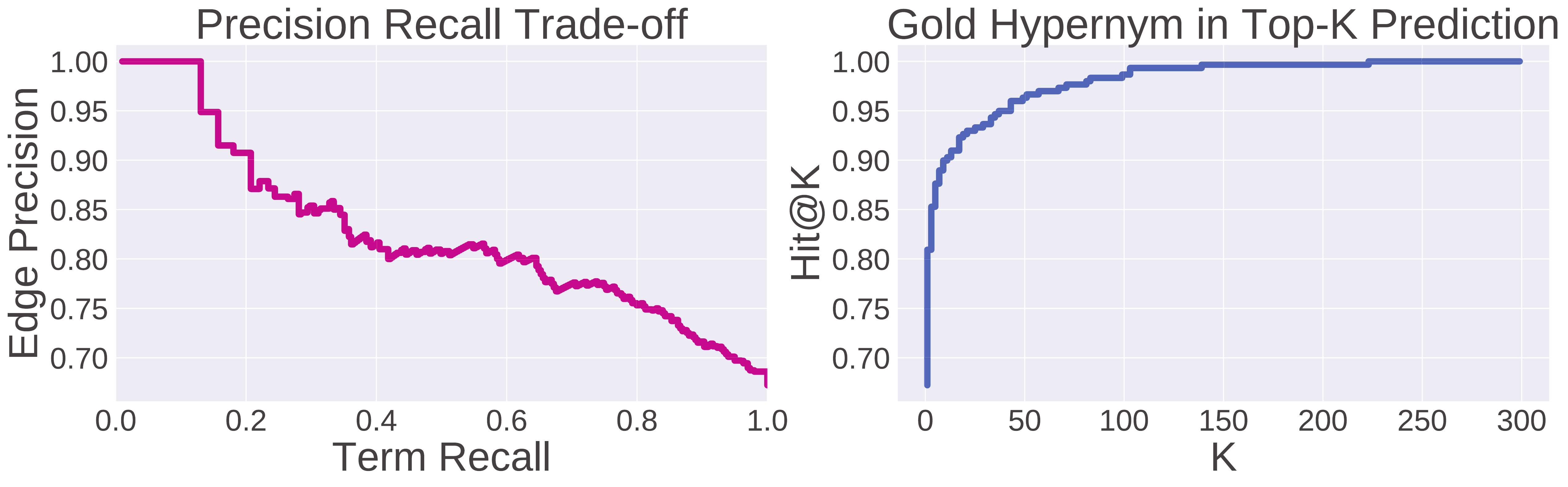}
    \vspace*{-.65cm}
    \caption{The precision recall trade-off \figleft\xspace and performance of term attachment in Hit@K \figright.}
    \label{fig:hit_pr}
       \vspace*{-.35cm}
\end{figure}

\noindent Precision recall trade-off answers the practical question in production ``how many terms can be attached if a specific precision of term attachment is required'', by filtering predictions with $\max_{v \in V} P(v' \mid v, \Theta) < c$, where $c \in [0, 1]$ is a thresholding constant.
As depicted in Fig~\ref{fig:hit_pr} \figleft, more than 15\% terms can be recalled and attached perfectly. Over 60\% terms can be recalled when an edge precision of 80\% is achieved.

We also analyze the performance changes if we relax term attachment to top-k predictions rather than top-1 prediction (as measured in Edge-F1).
We observe that more than 95\% gold hypernyms of the query terms are in the top-50 predictions (Fig~\ref{fig:hit_pr} \figright), which indicates that \ours generally ranks the gold hypernyms higher than other candidates even if its top-1 prediction is incorrect.
Note that the results in all the previous experiments are regarding term recall equal to 1 (all extracted terms are attached) and Hit@1 (whether the top-1 prediction is correct).

\subsubsection{Case Studies}

\noindent We inspect correct and incorrect term attachments to better understand model behavior and the contribution of each type of representation.
As shown in Table~\ref{table:expand_example}, \ours successfully predicts ``Fresh Cut Flowers'' as the hypernym of ``Fresh Cut Carnations'', where lexical representation possibly helps with the matching of ``Fresh Cut'' and semantic representation identifies ``Carnations'' is a type of ``Flowers''.
When lexical clues are lacking, \ours can still use semantic representation to recognize that ``Tilapia'' is a type of ``Fresh Fish''.
Without \gnnAware representation, \ours detects that ``Bock Beer'' is closely related to ``Beer'' and ``Ales''. By adding the \gnnAware representation \ours could narrow down its prediction to the more specific type ``Lager \& Pilsner Beers''.
For the wrongly attached terms, \ours is unconfident in distinguishing ``Russet Potatoes'' from ``Fingerling \& Baby Potatoes'', which is undoubtfully a harder case and requires deeper semantic understanding.  \ours also detects a potential error in the core taxonomy itself as we found the siblings of ``Pinto Beans'' all start with ``Dried'', which might have confused \ours during training.

\begin{table}[ht]
\footnotesize
    \caption{Case studies of term attachment. Correct and incorrect cases are marked in green and red, respectively.} 
    \label{table:expand_example}
        \vspace*{-.3cm}
    \centering
    \scalebox{.87}{
    \begin{tabular}{p{2.2cm}p{2.cm}p{4.5cm}}
        \toprule
         \textbf{Query Term} & \textbf{Gold Hypernym} & \textbf{Top-3 Predictions}\\
         \midrule
        \colorG{fresh cut carnations} & fresh cut flowers & \colorB{fresh cut flowers}, fresh cut root vegetables, fresh cut \& packaged fruits\\
        \midrule
        \colorG{tilapia} & fresh fish & \colorB{fresh fish}, liquor \& spirits, fresh shellfish \\
        \midrule
        \colorG{bock beers} & lager \& pilsner beers & \textbf{W/O \gnnAware representation:} ales, beer, tea beverages \newline  \textbf{Full Model:} \colorB{lager \& pilsner beers}, porter \& stout beers, tea beverages\\
        \midrule
        \colorR{fresh russet potatoes} & fresh potatoes \& yams & fresh fingerlings \& baby potatoes, fresh root vegetables, fresh herbs \\
        \midrule
        \colorR{pinto beans} & dried beans &  canned beans, fresh peas \& beans, single herbs \& spices \\
        \bottomrule
    \end{tabular}
    }
     \vspace*{-.25cm}
\end{table}

\section{Related Work}
\start{Taxonomy Construction.}
\cite{shen2012graph} proposed a graph-based approach to attach new concepts to Wikipedia categories.
\cite{jurgens2015reserating} enriched WordNet by finding terms from Wiktionary and attaching them via rule-based patterns.
A concurrent study \cite{shen2020taxoexpan} also leverages GNNs for taxonomy enrichment.
However, the features used in prior methods~\cite{shen2012graph,jurgens2015reserating,shen2018hiexpan,liu2019user} are designed either for specific taxonomies or for text-rich domains whereas \ours is robust to short texts and generally applicable to various online domains.
For \taxos, \cite{wang2014hierarchical} extracted terms from search queries and formulated taxonomy enrichment as a hierarchical Dirichlet process, where each node is represented by a uni-gram language model of its associated items.
\cite{liu2019user} used patterns to extract new concepts in the queries and online news, and built a topic-concept-instance taxonomy with human-labeled training set.
In contrast, \ours is self-supervised and utilizes heterogeneous sources of information.

\start{Term Extraction.}
\cite{kozareva2010semi} used Hearst patterns~\cite{hearst1992automatic} to extract new terms from the web pages.
\cite{shen2018hiexpan,zhang2018taxogen} used AutoPhrase~\cite{shang2018automated} to extract keyphrases in general-purpose corpora.
These methods, however, are inapplicable or ineffective for short texts like item titles.
On the other hand, many prior studies~\cite{bansal2014structured,panchenko2016taxi,jurgens2016semeval,bordea2016semeval,mao2018end} made a somewhat unrealistic assumption that one clean vocabulary is given as input and focused primarily on hypernymy detection.
Another plausible alternative is to treat entire user queries as terms rather than perform term extraction~\cite{wang2014hierarchical}, which results in a low recall.
In contrast, \ours employs a sequence labeling model designed for term extraction from online domains with self-supervision.

\start{Hypernymy Detection.}
Pattern-based hypernymy detection methods~\cite{hearst1992automatic,snow2005learning,kozareva2010semi,panchenko2016taxi,nakashole2012patty} consider the lexico-syntactic patterns between the co-occurrences of term pairs. They achieve high precision in text-rich domains but suffer from low recall and also generate false positives in domains like e-commerce.
Distributional methods~\cite{fu2014learning,rimell2014distributional,yu2015learning,tuan2016learning,shwartz2016improving} utilize the contexts of each term for semantic representation learning.
Unsupervised measures such as symmetric similarity~\cite{lin1998information} and distributional inclusion hypothesis~\cite{weeds2004characterising,chang2017unsupervised} are also proposed, but not directly applicable to \taxo enrichment as there are \textit{grouping nodes} in the \taxos like ``Dairy, Cheese \& Eggs'' and ``Flowers \& Gifts'' curated by taxonomists for business purposes, which might not exist in any form of text corpora.
In \ours, heterogeneous sources of signals are leveraged to tackle the lack of text corpora and grouping nodes are captured by head word semantics.

\section{Conclusion}
In this paper, we present a self-supervised end-to-end framework for \taxo enrichment that considers heterogeneous sources of representation and does not involve additional human effort other than the existing core taxonomies to be enriched.
We conduct extensive experiments on real-world taxonomies used in production and show that the proposed framework consistently outperforms state-of-the-art methods by a large margin under both automatic and human evaluations.
In the future, we will explore the feasibility of joint learning of \taxo enrichment and downstream applications such as recommendation and search.

\section*{Acknowledgment}
We thank Jiaming Shen, Jun Ma, Haw-Shiuan Chang, Giannis Karamanolakis, Colin Lockard, Qi Zhu, and Johanna Umana for the valuable discussions and feedback.

\bibliography{Taxo-Enrich}
\bibliographystyle{ACM-Reference-Format}

\newpage
\appendix
\section{Reproducibility}
\label{app:repro}

\subsection{Implementation Details}
We use fastText~\cite{bojanowski2017enriching} as the word embedding $\vw_v$ in order to capture sub-word information.
$\vw_v$ is fixed to avoid semantic shift for the inference of unseen terms $V'$.
All the terms $v$ are lowercased and concatenated with underscores if there are multiple words.

For \gnnAware representation, we use the Deep Graph Library\footnote{https://www.dgl.ai/} for the implementation of GNN-based models. The number of layers $L$ is set to 2 (two-hop neighbors are considered) and the normalization constant $c_{v, r}$ is 1.
We sample $N=5$ neighbors instead of using all the neighbors $N(v)$ for information aggregation.
Node embedding $\vh^{l}_v$ is of size 300, also initialized by the fastText embedding.
$\mW^l$ is of size 300 $\times$ 300.
All of the lexical string-level features $\mL_i(v, v')$ are randomly initialized and have an embedding size of 10.
$\mW_1$ and $\mW_2$ are of size 1 $\times$ 100 and 100 $\times\ |\mR(v, v')|$, respectively.

For \textit{relationship measure} involving the output of the GNNs, $s(\vv, \vv')$ measures the $\normlone$ Norm, $\normltwo$ Norm, and cosine similarity between the embeddings of a term pair.
For that between head words, $s(\vv, \vv')$ measures the cosine similarity of their corresponding word embeddings.
We use Adam as the optimizer with initial learning rate 1e-4 and choose the best performing model according to the development set.

\subsection{Data Availability}
The taxonomies used in the experiments are available online.
The taxonomies on Amazon.com can be obtained by scraping public webpages (refer to Fig.~\ref{fig:coffee_department}) or registering as a vendor.
The Yelp taxonomy is accessible at \url{www.yelp.com/developers/documentation/v3/all_category_list}.
The user behavior logs are mainly used for the GNN-based \gnnAware representation learning.
We use five-month user behavior logs in the target domain on Amazon.com and do not leverage user behavior on Yelp.com due to accessibility.
All of the experiments and framework analysis without the \gnnAware representation can be reproduced.

While each taxonomy used in the experiments is seemingly ``small'', it is \textit{real} and quite big (2K+ nodes) for a single domain (Grocery). \ours is easily applicable to other domains (Home, Electronics, ...), which also contain thousands of categories; they collectively form a taxonomy of 30K+ nodes. There are also datasets with similar or smaller size (\eg, 187 to 1386 nodes in \cite{zhang2016learning}, and 50 to 100 nodes in \cite{mao2018end, bansal2014structured}).
Our setup is more general than SemEval-2016 Task 14 \cite{jurgens2016semeval} and can be simplified to it if we ignore user behavior.

\section{Baseline}
\label{app:baseline}

\subsection{Baseline in Term Extraction}
NP chunking is one of the popular choices for general-purpose entity extraction~\cite{ren2015clustype}. We conduct NP chunking via spaCy\footnote{https://spacy.io/}, which performs dependency parsing on the sentences and finds noun phrases in them.
    No task-specific input (\eg, a list of positive terms) is required or supported for the training of NP chunking.
    Post-processing, including removing the terms containing punctuation or digits, is used to filter clearly invalid terms from the results.
    
AutoPhrase~\cite{shang2018automated} is the state-of-the-art phrase mining method widely used in previous taxonomy construction approaches~\cite{shen2018hiexpan,zhang2018taxogen}. 
    We replace the built-in Wikipedia phrase list that AutoPhrase uses for distant supervision with the terms $V$ on the core taxonomy $T$, as we find that it performs better than appending $V$ to the Wikipedia phrase list.
    Note that AutoPhrase uses exactly the same resources as \ours for distant supervision.
    In terms of methodology, AutoPhrase~\cite{shang2018automated} focuses more on the corpus-level statistics and performs phrasal segmentation instead of sequence labeling.

Pattern-based methods~\cite{liu2019user} are ineffective in our scenario due to the lack of clear patterns in the item profiles and user queries.
The classification-based method in~\cite{wang2014hierarchical} only works for term extraction from queries and also performs poorly, possibly because~\citet{wang2014hierarchical} treats the whole query as one term, which results in low recall.

\subsection{Baseline in Term Attachment}
HiDir~\cite{wang2014hierarchical} conducts hypernymy detection with the assumption that the representation of a taxonomy node is the same as its parent node, where one node is represented by its associated items.

Since we do not have \textit{description sentences} as in SemEval-2016 Task 14~\cite{jurgens2016semeval}, most of MSejrKu's features are inapplicable and we thus replace its features with those used in \ours except for the \gnnAware representation. we found that such changes improve MSejrKu's performance.

Substr~\cite{bordea2016semeval} is the sub-string baseline used in SemEval-2016 Task 13~\cite{bordea2016semeval}. It is shown that Substr, which regards A as B's hypernym if A is a substring of B (\eg, ``Coffee'' and ``Ground Coffee''), outperforms most of the systems in the SemEval-2016 competition on automatic taxonomy construction~\cite{bordea2016semeval}.

I2T matches $v'$ with item titles and finds the taxonomy nodes these items $I_{v'}$ are associated with. The final prediction of I2T is made by majority voting of the associated nodes where the items $I_{v'}$ are assigned to.

\section{Crowdsourcing}
\label{app:crowdsourcing}

Crowdsourcing is a reasonable choice in our scenario because our terms are common words (\eg, ``coffee'' and ``black coffee'') without complicated relations that require domain expertise (\eg, ``sennenhunde'' and ``entlebucher'' in WordNet).

\subsection{Crowdsourcing for Open-world Term Extraction Evaluation}
For crowdsourcing evaluation in term extraction, each item is assigned to 5 workers and only item types labeled by at least 2 workers are considered valid.
Links to the corresponding pages on Amazon.com are also provided to the workers.

\subsection{Crowdsourcing for End-to-End Evaluation}
We use crowdsourcing on MTurk for the end-to-end evaluation since expert evaluation is difficult to scale.
We assign each term to 4 MTurk workers for evaluation.
One critical problem of using crowdsourcing for end-to-end evaluation is that we can not ask MTurk workers to find the hypernym of each new term directly, as there are thousands of terms $V$ on the core taxonomy $T$ while the workers are unfamiliar with the taxonomy structure.
Alternatively, we ask the workers to verify the precision of \ours: we provide the predicted hypernym of \ours and the ancestors along the predicted path to the root as the candidates. 
We also include in the candidates one sibling term at each level along the predicted path. The workers are required to select the hypernym(s) of the query term $v'$ among the provided candidates.
In this way, we can estimate how accurate the term attachment is from the average customers' perspective.
Two other options, ``\textit{The query term is not a valid type}'' and ``\textit{None of above are hypernyms of the query term}'', are also provided, which corresponds to error types (A) and (B), respectively (refer to Sec.~\ref{sec:e2e_exp}).

\end{document}